\documentclass[11pt,a4paper]{article}
\usepackage[hyperref]{emnlp2020}
\usepackage{times}
\usepackage{latexsym}

\usepackage{microtype}

\usepackage{times}
\usepackage{url}
\usepackage{latexsym}
\usepackage{booktabs}
\usepackage{graphicx}

\usepackage{multirow}
\usepackage{amsmath}
\usepackage{interval}
\usepackage{amssymb}
\usepackage{float}

\usepackage{enumitem}
\usepackage[utf8]{inputenc}

\aclfinalcopy 
\def\aclpaperid{***} 

\title{ISAAQ - Mastering Textbook Questions with Pre-trained Transformers and Bottom-Up and Top-Down Attention}

\author{Jose Manuel Gomez-Perez \\
  Expert.AI Research Lab \\
  10 Prof. Waksman \\
  28036 Madrid, Spain \\
  \texttt{jmgomez@expertsystem.com} \\\And
  Raul Ortega \\
  Expert.AI Research Lab \\
  10 Prof. Waksman \\
  28036 Madrid, Spain \\
  \texttt{rortega@expertsystem.com} \\}

\date{}

\begin{document}
\maketitle
\begin{abstract}Textbook Question Answering is a complex task in the intersection of Machine Comprehension and Visual Question Answering that requires reasoning with multimodal information from text and diagrams. For the first time, this paper taps on the potential of transformer language models and bottom-up and top-down attention to tackle the language and visual understanding challenges this task entails. Rather than training a language-visual transformer from scratch we rely on pre-trained transformers, fine-tuning and ensembling. We add bottom-up and top-down attention to identify regions of interest corresponding to diagram constituents and their relationships, improving the selection of relevant visual information for each question and answer options. Our system ISAAQ reports unprecedented success in all TQA question types, with accuracies of 81.36\%, 71.11\% and 55.12\% on true/false, text-only and diagram multiple choice questions. ISAAQ also demonstrates its broad applicability, obtaining state-of-the-art results in other demanding datasets.
\end{abstract}

\section{Introduction}
\label{sec:intro}

Within NLP, machine understanding of textbooks is one of the grand AI challenges. As originally put by~\cite{Reddy_1988}: "Reading a chapter in a college freshman text (say physics or accounting) and answering the questions at the end of the chapter is a hard (AI) problem that requires advances in vision, language, problem-solving, and learning theory.". Towards such goals, the Textbook Question Answering (TQA) dataset presented in~\cite{Kembhavi2017AreYS} offers an excellent testbed. Drawn from middle school science curricula, it describes fairly complex phenomena through a combination of text and diagrams. Answering questions may therefore involve text, diagrams or both, and require information from multiple sentences and/or diagrams in long textbook lessons. 

Another characteristic of the TQA dataset that makes it rather unique and challenging is that questions often involve reasoning beyond methods based on co-occurrence analysis or simple look-up. TQA requires parsing information from different sentences, dealing with qualitative and quantitative information (\textit{high frequency} vs. \textit{20,000Hz}), and relating text or visual information with the question. Solving the TQA task also requires dealing with language about negation, conjunction, polarity or commonsense. On the visual side, TQA is rich with diagrams that describe potentially complex concepts, such as photosynthesis, the trophic chain, and the cycle of water, which are hard to represent as a single natural image. Quite on the contrary, diagrams contain simpler constituents and relationships between them, whose semantics needs to be captured in order to answer the questions.

Despite recent work, overall progress in the TQA dataset has been rather limited until now, suggesting that language and diagram understanding challenges like the ones above-mentioned are still far from solved. In this paper we address such limitations by building on the success of two recent developments in natural language processing and vision-and-language reasoning: large-scale, pre-trained language models and bottom-up and top-down (BUTD) attention. We demonstrate that, compared to previous approaches, transformer-based language models like BERT and RoBERTa can significantly contribute to increase the language understanding and reasoning capabilities required to answer TQA questions. We also show that BUTD attention, originally proposed for tasks like image captioning and visual question answering with natural images, can be effectively adapted to propose regions of interest in the diagram that are relevant for the question in hand, enabling the identification of diagram constituents and their relationships. The main contributions of this paper are the following: 
\begin{itemize}[noitemsep]
    \item For the first time, we apply transformers to language understanding in TQA, which involves fine-tuning of pre-trained transformers and ensembling.
    \item Based on BUTD attention we detect diagram constituents and their relationships, and link them to the question, its relevant background and answer options.
    \item We study the language and visual understanding capabilities of our approach, including several ablations, and demonstrate its robustness and broader applicability.
    \item We present ISAAQ (Intelligent System for Automatically Answering Textbook Questions), which implements our approach.
\end{itemize}

The remainder of the paper is structured as follows. Section~\ref{sec:rel} describes related work. Section~\ref{sec:pre} introduces the notation that will be used throughout the paper. We present our method in section~\ref{sec:method}, including an overview of the overall model, its main building blocks (background information retrievers, solvers, ensemble), and their interplay. We focus on each TQA question type, i.e. true/false and text and diagram multiple-choice (MC) questions, as sub-tasks of the main TQA task and propose specific solvers for each case, based on pre-trained transformers, fine-tuning and BUTD attention. In section~\ref{sec:exp}, we present our experimental results, including an ablation study focused on understanding the specific contribution of each solver and their components. We also analyze the robustness of our model, its ability to generalize to other datasets, and its reasoning abilities. Finally, section~\ref{sec:qual} illustrates the impact of the different techniques used in ISAAQ to address diagram MC questions in TQA.


\section{Related work}
\label{sec:rel}

In~\cite{Kembhavi2016ADI} several TQA baselines were proposed that were based on Machine Comprehension (MC) models like BiDAF~\cite{Seo2017BidirectionalAF} and MemoryNet~\cite{Weston2014MemoryN}, as well as Visual Question Answering (VQA)~\cite{Antol2015VQAVQ} and diagram parsing algorithms like DsDP-net~\cite{Kembhavi2016ADI}. Their results were rather modest (50.4, 32.9, and 31.3 in true/false, text and diagram MC questions), suggesting that existing MC/VQA methods would not suffice for the TQA dataset. Indeed, diagram questions entail greater complexity than dealing with natural images, as shown in~\cite{gomezlre2019}, where we beat the TQA baselines using visual and language information extracted from the correspondence between figures and captions in scientific literature enriched with lexico-semantic information from a knowledge graph~\cite{Denaux2019Vecsigrafo}. By contrast,~\cite{li2018} focused on finding contradictions between the candidate answers and their corresponding context while~\cite{Kim2019TextbookQA} applied graph convolutional networks on text and diagrams to represent relevant question background information as a unified graph.

The field of NLP has advanced substantially with the advent of large-scale language models such as ELMo~\cite{Peters}, ULMFit~\cite{howard2018ulmfit}, GPT~\cite{radford2018OpenAITransformer}, BERT~\cite{Devlin2018}, and RoBERTa~\cite{Liu2019RoBERTaAR}. Using large amounts of text, e.g. BERT was trained on Wikipedia plus the Google Book Corpus of 10,000 books~\cite{10.1109/ICCV.2015.11}, they are trained to learn various language prediction tasks such as guessing a missing word or the next sentence. Language models and particularly transformers have been used in question answering, as illustrated by the success of the Aristo system~\cite{Clark2019FromT} in standard science tests. Transformers have also proved their worth as soft reasoners~\cite{Clark2020TransformersAS}, exhibiting capabilities for natural language inference. Furthermore, whilst learning linguistic information, transformers have shown to capture semantic knowledge and general understanding of the world from the training text~\cite{petroni-etal-2019-language}, including a notion of commonsense that can be useful in question answering. Our approach is the first to leverage the language understanding and reasoning capabilities of existing transformer language models for TQA. 

Focused on natural images, some vision-and-language reasoning systems are also adopting transformer architectures at their backbone. VL-BERT~\cite{su2019vl} and LXMERT~\cite{tan-bansal-2019-lxmert} pre-train large-scale transformers that capture both visual concepts and language semantics, as well as cross-modal information. Pre-training is done via several tasks, like masked language modeling, masked object prediction, cross-modality matching, and image question answering, on large-scale text and visual datasets, like Conceptual Captions~\cite{sharma-etal-2018-conceptual}, Google Book corpus, MS COCO~\cite{Lin2014MicrosoftCC} or Visual Genome~\cite{10.1007/s11263-016-0981-7}, requiring considerable compute. By contrast, our approach is much more frugal. We fine-tune pre-trained existing transformers like BERT and RoBERTa and count on a much more limited variety of datasets focused on diagrams, like AI2D~\cite{Kembhavi2016ADI}. Like LXMERT, we extend region-of-interest (RoI) features with object positional embeddings. Finally, we apply bottom-up and top-down attention~\cite{8578734} to focus on the most relevant diagram regions for each question.   

\section{Preliminaries and notation}
\label{sec:pre}

\subsection{The TQA dataset}
The TQA dataset~\cite{Kembhavi2017AreYS} comprises 1,076 lessons from Life Science, Earth Science and Physical Science textbooks, with 78,338 sentences and 3,455 diagrams, distributed in 5,400 true/false questions, 8,293 text MC, and 12,567 diagram MC questions. The dataset is split in training, validation and test sets (table~\ref{table:datasets}), which are disjoint at the lesson level. Thus, our model will often need to answer questions it was not trained for, which entails additional challenges to generalize beyond the training set (section~\ref{sec:exp}). TQA questions are long compared to VQA, with a mode of 8 vs. 5 words per question. Almost 85\% of the questions are \textit{what}, \textit{how} or \textit{which} wh- questions. Another 10\% is formulated assertively, bringing additional language understanding complexity. Most (80\%) text MC questions can be answered with information from one or several sentences in a paragraph. The rest may require multiple paragraphs and lessons as well as external knowledge. Over 40\% diagram MC questions require complex diagram parsing, only 2\% can be answered with an OCR. 

\subsection{Notation}
We divide the TQA task in three sub-tasks, one per question type. The solvers addressing each sub-task are denoted as $TF_m$ (true/false questions), $TMC_m$ (text MC) and $DMC_m$ (diagram MC). 
Suffix $m$ indicates the method used for background retrieval.

$\forall l \in L$, let $ls_i$ be each sentence in lesson $l$, where $L$ is the set of lessons in the dataset. We apply BERT-style transformers~\cite{Devlin2018} to MC questions, treating the task as multiple choice classification. Given a question $q \in l$ with answer options $a_i$ and background knowledge $K$, we pass the following sequence $s$ to the transformer:
\begin{equation}
\label{eq:seq}
\begin{split}
seq(K,QA_i)&= [CLS]K[SEP]QA_i[SEP]
\end{split}
\end{equation}
with $QA_i=[q,a_i]$. Similarly, for true/false questions we explore the relation between a question $q$ and a sentence $ls$. Overloading the previous method, $s$ is obtained as: 
\begin{equation}
\label{eq:seq2}
\begin{split}
seq(q,ls)&= [CLS]q[SEP]ls[SEP]
\end{split}
\end{equation}

A transformer $T$ will produce one vector for each token in $s$, including $[CLS]$, whose vector we denote as $T_{[CLS]}(s)$, which we use as a pooled representation of the whole sequence. 



\section{Proposed Method}
\label{sec:method}
\begin{figure*}[t]
    \centering
    \includegraphics[width=0.9\textwidth]{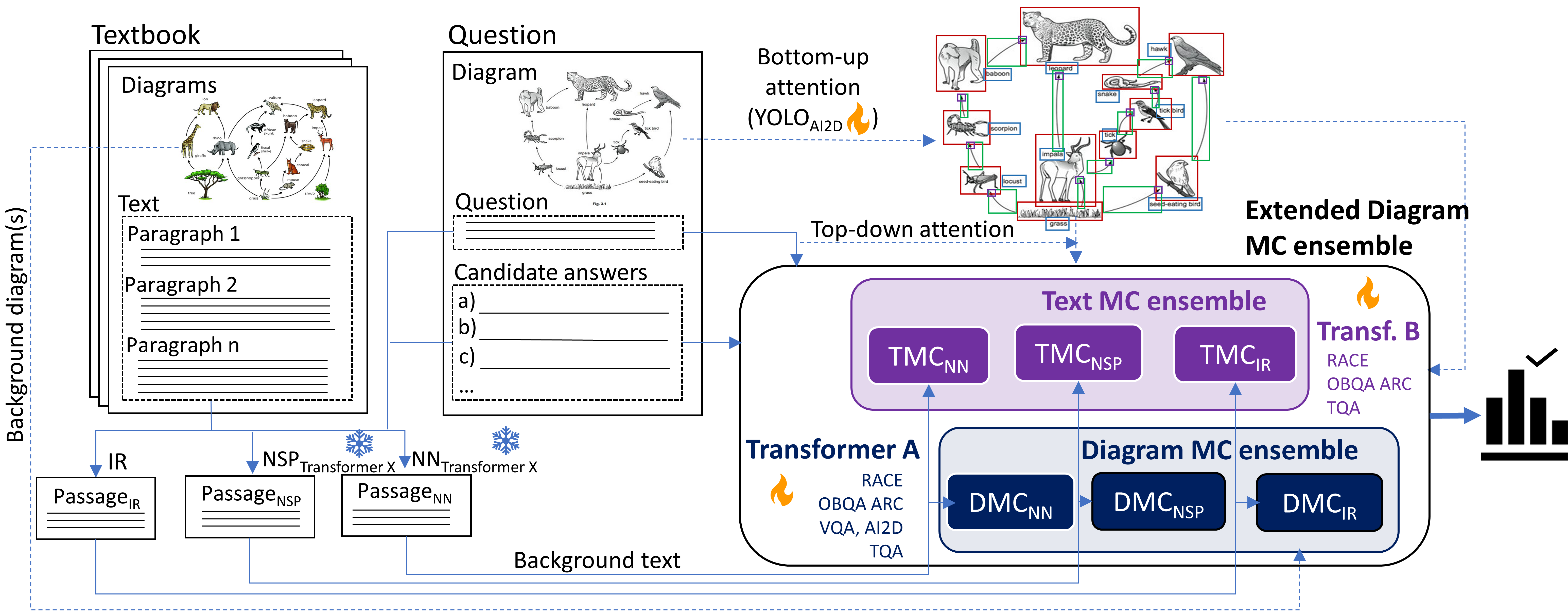}
    \caption{Overview of the proposed TQA model for text and diagram MC. Left: background retrieval stage. Right: Text and Diagram MC solvers are trained using transformers and BUTD. Flames indicate the underlying transformer is fine-tuned for the task at hand. Ice denotes the opposite. Dashed lines only apply for diagram MC.}
   \label{fig:model}
\end{figure*}

Figure~\ref{fig:model} shows our two-stage process to answer text and diagram MC questions. First, for each question we propose different retrievers to extract relevant language and visual background knowledge from the textbook. Note that we consider both approaches based on conventional information retrieval techniques and approaches that leverage transformers pre-trained on specific tasks.

During training, the retrieved background is provided along with the question and candidate answers to our solvers. Also during execution, providing the potential to “read” such knowledge and apply it to the question. We ensemble different solvers resulting from fine-tuning one or several transformers on a multiple choice classification task, which can be combined with others based e.g. on information retrieval. 

For text and diagram MC questions, each transformer-based solver results from training the MCC task on the text passages produced by one of the text retrieval methods. Since each text retriever produces a different but complementary dataset of background text passages, the resulting solvers also complement each other, motivating their combination as an ensemble. In addition, for the visual part we apply BUTD attention as shown in figure~\ref{fig:DMC}. For true/false questions we follow an analogous two-stage process, in this case fine-tuning our transformers on a text entailment task. 

\subsection{Background knowledge retrieval} 
\label{subsec:background}


Information retrieval techniques to obtain background information from the text are usually keyword-based and potentially oblivious of the different artifacts of language, such as morphological variations, conjugations, terms that may be semantically related to the question, synonyms, hypernyms or multi-word expressions, which are particularly frequent in the domains of the TQA dataset. To address such shortcomings, we extend classic information retrieval approaches with pre-trained models that leverage the language understanding capabilities of transformer language models. This results in three text background retrievers: 

\textbf{Information Retrieval (IR)}
The IR method searches the whole TQA dataset to see if question $q$ along with an answer option is explicitly stated in the corpus. 
For each answer option $a_i$, we concatenate $q$ and $a_i$ and run the query against a search engine like ElasticSearch. Based on the search engine score, we take the top $n$ sentences ($n=10$) resulting from the query, where each sentence has at least one overlapping, non-stop word with $a_i$. This ensures that all sentences have some relevance to both $q$ and $a_i$, while maximizing recall. Then, we concatenate the selected sentences following their ranking to compose a text passage with the desired background knowledge.

\textbf{Next Sentence Prediction (NSP).} We implement this retriever by treating the task as next sentence prediction using a transformer $T$ with frozen weights. For each triple $(q,a_i,ls_j)$ we produce a sequence $s_{ij}=seq([q,a_i],ls_j)$, where $q$ is a question, $a_i$ one of its possible answers, and $ls_j$ a sentence in lesson $l$. We pass it to $T$ and take the probability that $ls_j$  can be semantically derived from the statement that $a_i$ is the answer of $q$, with label ${\tt isNext}$. Then, we rank the sentences based on such value, take the top $n$ sentences, and return the passage resulting from their concatenation.

\textbf{Nearest Neighbors (NN).} 
For each question and candidate answer pair $q,a_i$ and sentence $ls_j$ in lesson $l$, we obtain their vector representations $C_{i}=T_{[CLS]}([q,a_i])$ and $C_{j}=T_{[CLS]}(ls_j)$. We calculate the cosine similarity between them, take the top $n$ sentences based on their similarity score, and concatenate them as a single paragraph. 

\textbf{Diagram retrieval.} In addition to text, we also retrieve background diagrams. To this purpose, we pass the question and lesson diagrams through a ResNet-101 network pre-trained on ImageNet~\cite{imagenet_cvpr09}. We calculate the cosine similarity between the resulting features and select the lesson diagram closest to the question diagram.


\subsection{Solvers}
\label{sec:solvers}
The ISAAQ solvers result from the combination of three main components: i) the specific models used to address each TQA question type as a particular sub-task within the overall model, ii) the underlying transformer language model, and iii) the background information associated to each question used to train the solver. Here we focus on the first of such components for the different types of questions in the TQA dataset: true/false, text, and diagram MC questions. 

\textbf{True/False Questions.} We address true/false questions as an entailment task, where question $q$ corresponds to the hypothesis and the premise is a sentence $ls_i$, with $q,ls_i \in$ lesson $l$. Such task is modeled as sequence classification, using a pre-trained transformer $T$. For each $ls_i$, sequence $s_i=seq(q,ls_i)$ is passed to the sequence classification model, obtaining a $2-d$ logit vector. The answer, with possible labels ${\tt true}$ or ${\tt false}$, is computed as the output of a binary classifier, trained by minimizing the negative log-likelihood of the correct answer produced by a softmax layer. 

\textbf{Text MC Questions.} This solver aims to select the answer to question $q$ amongst several answer options $a_i$, where the retrieved background knowledge for $q$ and $a_i$ is a passage $p$. To this purpose we use a pre-trained transformer $T$ to implement a multiple choice classification model. For each $a_i$, we pass the input sequence $s_i=seq(p,[q,a_i])$ and obtain and $N-d$ logit vector, with $N$ the number of answer options. The answer to $q$ is the output of a multi-class classifier, also trained by minimizing the negative log-likelihood of the correct answer.




\textbf{Diagram MC Questions}

\begin{figure*}[t]
    \centering
    \includegraphics[width=0.9\textwidth]{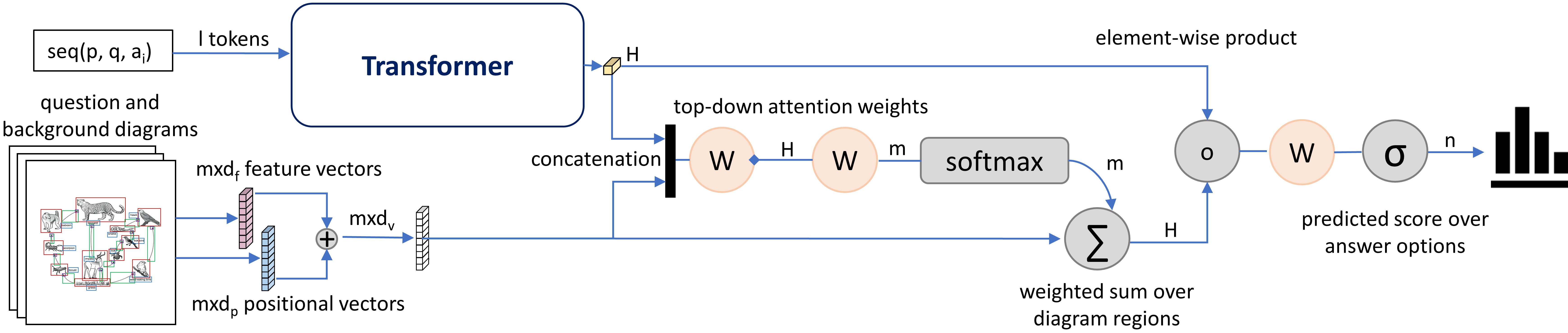}
    \caption{Overview of solver architecture for diagram questions, including BUTD attention.}
   \label{fig:DMC}
\end{figure*}

Since diagram questions involve both text and diagrams, we need to address both branches in our model and merge them in order to answer the question based on both text and visual information (see figure~\ref{fig:DMC}). To encode the text part, we follow the same approach as with text MC questions. For the visual part, instead of using the feature map of the diagram produced by a convolutional neural network, we apply BUTD attention~\cite{8578734} to take the features of the regions of interest (RoI) detected bottom-up in the diagram and then apply top-down attention on the question. 

Each RoI $r_j \in \{r_1 \ldots r_m\}$ is represented by two vectors: a 
visual feature vector $f_j$ with dimensionality $d_f=1000$ and a positional vector $p_j$ with dimensionality $d_p$ containing 4 bounding box coordinates. 
In contrast to directly using the feature vector $f_j$ as in~\cite{8578734} and in line with other work like~\cite{tan-bansal-2019-lxmert}, we learn an embedding $v_j$ of dimensionality $d_v=1024$:
\begin{equation}
\begin{split}
\hat{f_j}&=LayerNorm(W_F,f_j+b_f) \\
\hat{p_j}&=LayerNorm(W_P,p_j+b_P) \\
v_j&=(\hat{f_j}+\hat{p_j})/2
\end{split}
\end{equation}

To extract RoIs and their positional information we fine-tune YOLO~\cite{Redmon2015YouOL} on AI2D~\cite{Kembhavi2016ADI}, a dataset with TQA-style diagrams annotated with position and region type. The visual features $v_j$ of each RoI detected bottom-up by YOLO are also made to attend top-down to the representation of question $q$, its background knowledge $p$ and answer options $a_i$, produced by a transformer language model $T$ (figure~\ref{fig:DMC}). $\forall a_i$ we provide $T$ with input $s_i=seq(p,[q,a_i])$ and obtain $C_i=T_{[CLS]}(s) \in \mathbb{R}^{H}$, a pooled representation of $s_i$. We generate an unnormalized top-down attention weight $a_{ij}$ for each $m$ diagram RoI feature vector $v_j$ as:
\begin{equation}
a_{ij} = w_{a}^Tg_a([v_j,C_i])    
\end{equation}

where $w_{a} \in \mathbb{R}^{H}$ is a learned vector. We implement the learned non-linear transformation $g$ as a gated hyperbolic tangent activation~\cite{10.5555/3305381.3305478}. The normalized attention weight and attended image feature $\hat{v_i}$ for each option $a_i$ are:
\begin{equation}
\begin{split}
\alpha_{ij}&=softmax(a_{ij}) \\
\hat{v_i}&=\sum_{j=1}^{m}\alpha_{ij}v_j
\end{split}
\end{equation}

The distribution $\hat{y}$ over the possible outputs is:
\begin{equation}
\begin{split}
\hat{y}&=softmax(UW_u)    
\end{split}
\end{equation}
where $U \in \mathbb{R}^{NxH}$ is a matrix of $u_i$ vectors, with $N$ the number of answer options $a_i$, and $W_u \in \mathbb{R}^{H}$ 
a learned parameter vector. Each vector $u_i$ is a joint representation of the question and the diagram for answer option $a_i$, where $u_i=C_i \circ \hat{v_i}$ .


\subsection{Ensemble}
\label{subsec:ens}

The choice of a specific background retriever may have a significant impact in the overall performance of each solver after training. Transformer-based background retrieval methods have deeper language understanding capabilities than those based on classic information retrieval approaches. However, they are also more computationally demanding. This has implications in terms of the textbook range that each retriever can reasonably cover. To address such trade-off we use information retrieval methods to extract background sentences from the whole textbook, knowing they may not be as accurate, while transformer-based methods focus on the lesson of the question to be answered, which potentially contains more relevant information. 

We train our solvers using variations of the background knowledge provided by the different retrieval methods. Then, for each question type, we combine the resulting solvers in a single ensemble. Our ensemble algorithm is based on the two-step approach proposed in~\cite{Clark2019FromT} to produce a combined score in $\interval{0}{1}$. In the first step, each solver $s$ is calibrated by learning a logistic regression classifier from each answer option to a correct/incorrect label. Like~\cite{Clark2019FromT}, we also calibrate on the training set. The features for answer option $a_i$ include the raw score $s_i$ and its value across all question options, normalized with a softmax. This step returns a calibrated score per solver $s$ and option $a_i$. The second step uses the calibrated scores as the input to another logistic regression classifier whose output is the ensemble score for each $a_i$.




\section{Experiments and Results}
\label{sec:exp}

\subsection{Experimental settings}
Our approach is rather frugal in terms of hardware. All training and evaluation has been done on a single server with 32GB of RAM, 1TB SSD and a single GPU GeForce RTX 2080 Ti. ISAAQ solvers have been implemented\footnote{All models, source code, and libraries are available at \\ \url{https://github.com/expertailab/isaaq}} using the Transformers library\footnote{\url{https://huggingface.co/transformers}} and RoBERTa large. We apply Pareto to select maximum input sequences of 64 tokens for true/false questions and 180 for text and diagram MC. For text background retrieval we use a pre-trained BERT-base model
. We train each TQA sub-task during 4 epochs and pick the epoch with the best accuracy. We take Adam~\cite{Kingma2014AdamAM} with linearly-decayed learning-rate and warm-up as in~\cite{Devlin2018} and empirically select peak learning rates in the range $\interval{1e^{-6}}{5e^{-5}}$, with $1e^{-5}$ for true/false and text MC questions and $1e^{-6}$ for diagram MC. Similarly, we choose a dropout value of 0.1 at the exit of the transformer. Training time per epoch is 1' for true/false questions, 30' for text MC, and 60' for diagram MC.

For diagram encoding, we pass each RoI to a pre-trained ResNet-101~\cite{resnet2016} backbone. We have experimented with other visual models like VGG~\cite{DBLP:journals/corr/SimonyanZ14a} with similar results. Unlike ~\cite{tan-bansal-2019-lxmert} and~\cite{8578734}, who used Faster R-CNN~\cite{10.1109/TPAMI.2016.2577031} on natural images, we choose YOLO to extract RoIs from TQA diagrams. After fine-tuning on AI2D, YOLO outperformed Faster R-CNN with a test set accuracy of 81.2\% vs 79.22\%. Both results suggest around 20\% margin for additional improvement in RoI selection. We apply Pareto to fix the maximum number of regions to 32 and fine-tune YOLO on AI2D with standard parameters\footnote{\url{https://github.com/ultralytics/yolov3}} for 242 epochs and initial learning rate $1e^{-4}$.  

\subsection{Language and visual pre-training} 
We pre-train our text MC question solvers on several datasets (table~\ref{table:datasets}), including some not specific of science. The resulting fine-tuned transformer is also used to train the true/false solvers. We follow common practice in multi-step fine-tuning, with some variations in the usual order based on dataset size. First, we fine-tune on the training set of RACE~\cite{sun-etal-2019-improving}, a challenging set of English comprehension MC exams given in Chinese middle and high schools. Then we continue with the training sets of a collection of scientific MC question datasets: ARC~\cite{Clark2018ThinkYH}, both Easy and Challenge, and OpenBookQA~\cite{mihaylov-etal-2018-suit}. Finally, we fine-tune the result of the previous step on the TQA training set for text and diagram MC. Peak learning rates are $1e^{-6}$ for the first fine-tuning step and $1e^{-5}$ for the second.

We pre-train our diagram MC question solvers on the training sets of VQA abstract scenes and VQA, the latter being the largest visual resource available with support for MC questions and diagram-style images. The size of such datasets is still far from natural image datasets like Visual Genome or MS COCO. Also, note that AI2D is annotated for generic diagram constituents (blob, arrow, arrow head, text), i.e. it does not observe semantic visual categories like cloud, tree or mammal. Nor does AI2D annotate parts of diagram blobs, like the different layers of Earth or the organelles in a cell, which suggests further room for improvement. We train on VQA and AI2D with learning rate $1e^{-6}$ for 4 and 12 epochs, respectively.


\begin{table}[htbp]
\small
\centering
\begin{tabular}{@{}lcccc@{}}
\toprule
 & \multicolumn{3}{c}{Partition}   &  \\ \cmidrule(lr){2-4}
Dataset & Train & Dev & Test & Total \\ \midrule
RACE & 87,866 & 4,887 & 4,934 &  97,687\\ 
ARC-Easy & 2,251 & 570 & 2,376 & 5,197\\
ARC-Challenge & 1,119 & 299 & 1,172 & 2,590\\
OpenBookQA & 4,957 & 500 & 500 & 5,957 \\ \midrule
VQA (abs. scenes) & 60,000 & 30,000 & 60,000 & 150,000 \\
AI2D & 7,824 & 906 & 978 & 9,708 \\ \midrule
TQA & 15,154 & 5,309 & 5,797 & 26,260 \\
\bottomrule
\end{tabular}
\caption{Dataset partition sizes (\#questions).}
\label{table:datasets}
\end{table}



\subsection{Main results}

\begin{figure}[h]
    \centering
    \includegraphics[width=0.49\textwidth]{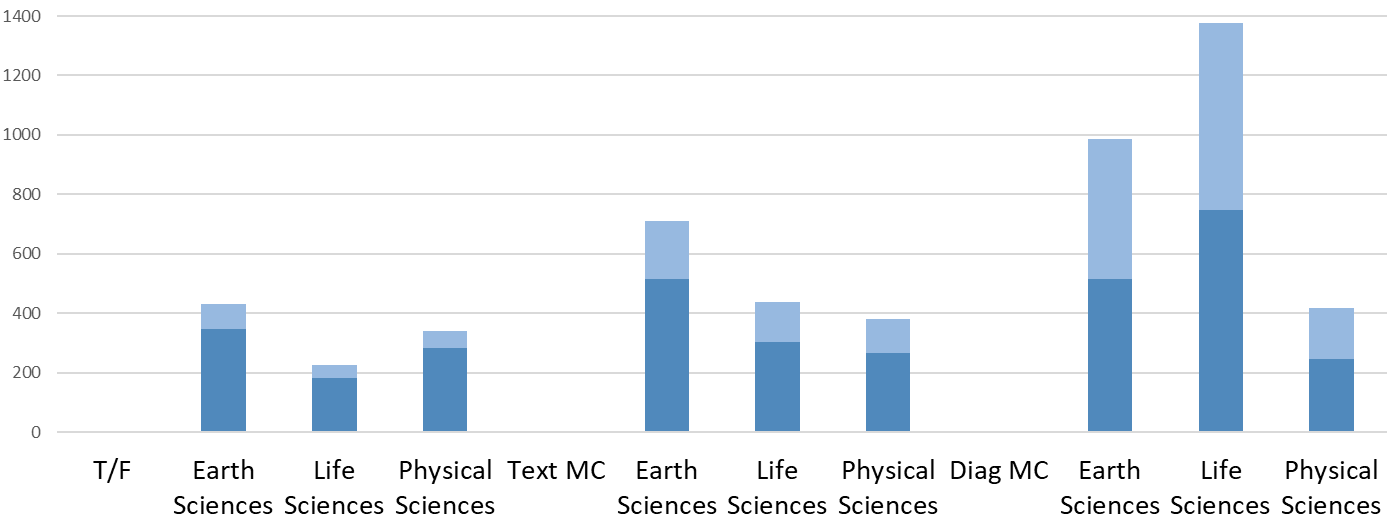}
    \caption{ISAAQ hits (darker) vs. misses (lighter).}
   \label{fig:hitsmisses}
\end{figure}

Table~\ref{table:results} shows the results (\% accuracies) obtained by ISAAQ in true/false, text and diagram MC. Figure~\ref{fig:hitsmisses} shows the ratio of correct vs. incorrectly answered questions per question type and subject matter. Results are very similar across all domains, with a slight preference for Physical sciences.

Since most previous approaches did not report on the test set, here we focus on validation. ISAAQ outperforms all other systems in all question types, establishing by a large margin a new SotA in TQA. 
We include a baseline using RoBERTa and a VQA model which, unlike ISAAQ, does not apply e.g. fine-tuning on related datasets or ensembling. This baseline outperforms all previous approaches, illustrating the benefits of applying transformers to this problem. However, ISAAQ clearly beats it in all question types, particularly in diagram MC, with a 14\% improvement  that demonstrates how ISAAQ successfully incorporates visual information into the transformer. On the other hand, our results also confirm the complexity of diagram MC compared to the other question types, suggesting future work in language-visual understanding. Finally, ISAAQ also obtains excellent results in all the datasets used for pre-training (table~\ref{table:datasets}), confirming that our approach is robust and generalizes well.

\begin{table*}[htbp]
\footnotesize
\centering
\begin{tabular}{lcccccc}
\toprule
\multicolumn{2}{c}{Model} & Text T/F & Text MC & Text All & Diagram MC & All \\ \midrule
Random & - & 50.10 & 22.88 & 33.62 & 24.96 & 29.08 \\
MemN+VQA & \multirow{3}{10em}{\cite{Kembhavi2017AreYS}} & 50.50 & 31.05 & 38.73 & 31.82 & 35.11 \\
MemN+DPG & & 50.50 & 30.98 & 38.69 & 32.83 & 35.62 \\
BiDAF+DPG & & 50.40 & 30.46 & 38.33 & 32.72 & 35.39 \\
FCC+Vecsigrafo & \cite{gomezlre2019} & - & 36.56 & - & 35.30 & - \\
IGMN & \cite{li2018} & 57.41 & 40.00 & 46.88 & 36.35 & 41.36 \\
f-GCN1+SSOC & \cite{Kim2019TextbookQA} & 62.73 & 49.54 & 54.75 & 37.61 & 45.77 \\ \midrule
RoBERTa+VQA & & 76.85 & 62.81 & 68.38 & 41.14 & 54.09\\
\textbf{ISAAQ} & & 81.36 & 71.11 & 75.16 & 55.12 & 64.66 \\
\bottomrule
\end{tabular}
\caption{ISAAQ performance and comparison (validation set) with previous SotA for the TQA dataset.}
\label{table:results}
\end{table*}


\begin{table}[h] 
\footnotesize
\centering
\begin{tabular}{lccc}
\toprule
Dataset & ISAAQ & \multicolumn{2}{l}{SotA} \\ \midrule
RACE & 71.63 & 90.90 &\cite{Shoeybi2019MegatronLMTM}\\
OBQA & 83.60 & 86.00 & \multirow{3}{9em}{\cite{Khashabi2020UnifiedQACF}} \\
ARC-Easy & 83.51 & 85.70 & \\
ARC-Cha. & 60.34 & 75.60 & \\ \midrule
VQA abs. & 64.75 & 74.37 & \cite{Teney2017GraphStructuredRF} \\
AI2D & 73.29 & 38.47 & \cite{Kembhavi2016ADI}\\
\bottomrule
\end{tabular}
\caption{ISAAQ vs. SotA in pre-training datasets (test).}
\label{table:datasets}
\end{table}

\begin{figure}[h]
    \centering
    \includegraphics[width=0.43\textwidth]{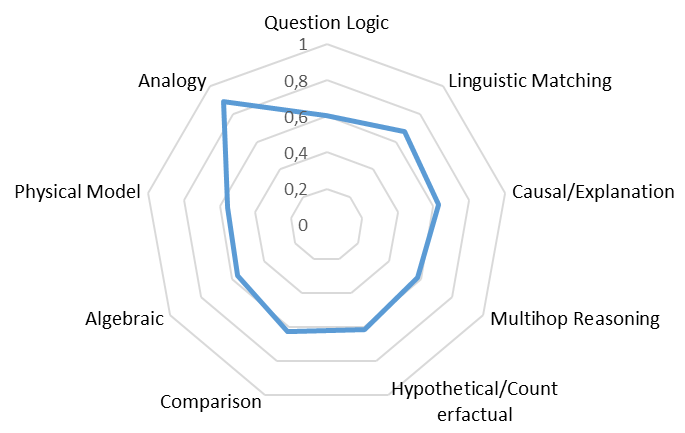}
    \caption{ISAAQ performance per reasoning type.}
   \label{fig:know-res}
\end{figure}

To obtain a deeper understanding of ISAAQ's reasoning ability we focus on ARC-Challenge, which only contains questions that neither retrieval nor co-occurrence methods can answer correctly. We run ISAAQ on a sample of 203 text MC questions manually annotated by~\cite{boratko-etal-2018-systematic} against 7 knowledge and 9 reasoning types. Since these questions were extracted from the ARC-Challenge training set, for this experiment we previously removed them from the pre-training of our model. Figure~\ref{fig:know-res} shows how the results we obtain for each reasoning type are in general in line with our overall results in the ARC-Challenge test set (60.34\%). We also notice an interesting spike in analogical reasoning, featured in~\cite{Kembhavi2017AreYS} as a key reasoning type in TQA, with 90\% accuracy. This is consistent with the findings reported by~\cite{Clark2019FromT,Clark2020TransformersAS} on the reasoning ability of transformer language models.  

\subsection{Ablation study}
We drill down on the results reported in table~\ref{table:results} in order to understand the contribution of each solver to the ISAAQ ensemble, both in the validation and test sets. Tables~\ref{table:TF},~\ref{table:textMC}, and~\ref{table:diagMC} show the individual results of such solvers for true/false, text, and diagram MC questions. For each sub-task, the differences between solvers result mainly from the background information used for training, which was produced by different retrievers (section~\ref{subsec:background}).
\begin{table}[htbp]
\footnotesize
\centering
\begin{tabular}{lccccc}
\toprule
Dataset & ${\tt TF_{IR}}$ & ${\tt TF_{NSP}}$ & ${\tt TF_{NN}}$ & \textbf{ISAAQ}\\
\midrule
validation & 78.26 & 76.25 & 79.16 & 81.36\\ 
test & 77.74 & 74.89 & 75.44 & 78.83 \\
\bottomrule
\end{tabular}
\caption{Results of each of our solvers and the overall ISAAQ model for TQA true/false questions. }
\label{table:TF}
\end{table}


In addition to RoBERTa large we experimented with BERT large, adding 3 more solvers to the true/false and MC ensembles. However, the added gain was extremely limited. Replacing BERT with RoBERTa large without pre-training on the datasets shown in table~\ref{table:datasets} had similar effect. Thus, we opted for the 3-way ensembles shown in the tables.
\begin{table}[htbp]
\footnotesize
\centering
\begin{tabular}{lccccc}
\toprule
Dataset & IR & ${\tt TMC_{IR}}$ & ${\tt TMC_{NSP}}$ & ${\tt TMC_{NN}}$ & \textbf{ISAAQ}\\
\midrule
validation & 47.91 & 67.52 & 68.63 & 64.64 & 71.11 \\ 
test & 48.31 & 68.94 & 67.19 & 65.31 & 72.06 \\
\bottomrule
\end{tabular}
\caption{Individual text MC solvers and ISAAQ. Note the large delta vs. IR solver baseline (also in table~\ref{table:diagMC}). Pre-training on RACE, OBQA, ARC-Easy/Challenge.}
\label{table:textMC}
\end{table}

Tables~\ref{table:textMC} and~\ref{table:diagMC} show how the solvers based on transformers clearly outperform an information retrieval baseline (IR), which is itself already on par with the former TQA SotA~\cite{Kim2019TextbookQA} for text MC questions and clearly better for diagram MC. Each of the true/false, text MC, and diagram MC solvers perform similarly in their respective question type sub-tasks. They are also complementary: in average, 33.41\% of the questions answered incorrectly by one of the solvers is correctly addressed by another. Such complementarity brings an extra performance boost by combining the different solvers in each sub-task as an ensemble. 
\begin{table}[h]
\footnotesize
\centering
\begin{tabular}{lccccc}
\toprule
Dataset & IR & ${\tt DMC_{IR}}$ & ${\tt DMC_{NSP}}$ & ${\tt DMC_{NN}}$ & \textbf{ISAAQ}\\
\midrule
validation & 39.12 & 53.83 & 52.14 & 51.28 & 55.12\\
test & 32.57 & 50.50 & 50.84 & 51.08 & 51.81\\
\bottomrule
\end{tabular}
\caption{Individual diagram MC solvers and ISAAQ. Pre-training on VQA abstract scenes and AI2D.}
\label{table:diagMC}
\end{table}

While the results obtained for text MC questions do not seem to depend on the specific split, for true/false and diagram MC questions we obtain clearly better results in the validation set compared to the test set. This emphasizes the heterogeneity of the TQA splits and how challenging it is to produce a model that generalizes well across them.

Looking at the incremental analysis of our diagram MC model in table~\ref{table:DMC}, visual information only enters into play once pre-training on VQA and AI2D is added, outperforming the text baseline. Additional background diagrams adds a little in test. BUTD attention does improve considerably in both validation and test, but not much more than using just bottom-up (BU) attention. The final ISAAQ model is a 6-way ensemble that combines the transformer-based solvers for text MC (table~\ref{table:textMC}) and diagram MC (table~\ref{table:diagMC}) and uses VQA$+$AI2D pre-training and BUTD. These results indicate interesting challenges yet to be addressed.





\begin{table}[htbp] 
\footnotesize
\centering
\begin{tabular}{lcc}
\toprule
Model & Val. acc. & Test acc. \\
\midrule
text (w/o pre-training) & 46.67 & 39.79\\
text & 53.22 & 46.82 \\ \midrule
text+visual (w/o pre-training) & 51.31 & 47.34 \\
text+visual & 53.54 & 51.32\\
text+visual+background diagram & 53.47 & 51.84 \\
text+visual+BU attention & 53.93 & 51.60  \\
text+visual+BUTD attention & 54.26 & 52.15\\
\midrule
\textbf{ISAAQ} & 55.12 & 51.81\\
\bottomrule
\end{tabular}
\caption{ISAAQ ablations for diagram MC.}
\label{table:DMC}
\end{table}

\section{Qualitative study}
\label{sec:qual}

\begin{table*}[t]
  \footnotesize
  \centering
  \begin{tabular}{cccc}
  \toprule
    Question & Diagram & Diagram MC+BU & Diagram MC+BUTD \\ \hline
    \begin{minipage}{.2\textwidth}
    \vspace{2mm}
      Which of the following layers comprises mineral particles?
    \vspace{-3mm}
    \begin{enumerate}[noitemsep]
    \item bedrock
    \item subsoil
    \item surface layers
    \item topsoil \checkmark
    \end{enumerate}
    \end{minipage} &
    \begin{minipage}{.23\textwidth}
      \includegraphics[width=\linewidth]{question831.png}
    \end{minipage} &
    \begin{minipage}{.23\textwidth}
      \includegraphics[width=\linewidth]{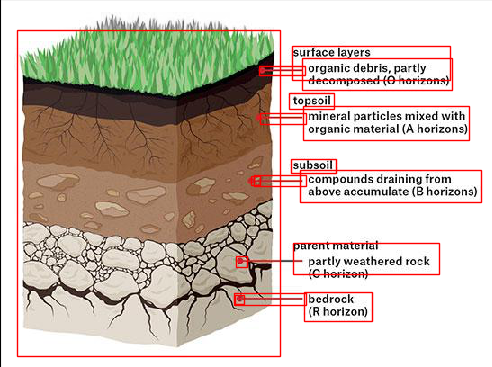}
    \end{minipage} &
    \begin{minipage}{.24\textwidth}
      \includegraphics[width=\linewidth]{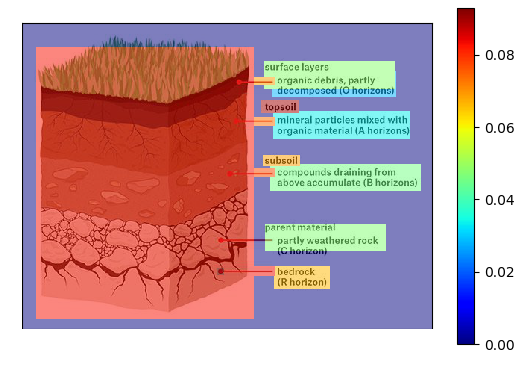}
    \end{minipage} 
    \\ 
    \begin{minipage}{.21\textwidth}
       Which phase is shown in the picture below?
    \vspace{-3mm}
    \begin{enumerate}[noitemsep]
    \item mitosis \checkmark
    \item prophase
    \item interphase
    \item mitotic 
    \end{enumerate}
    \end{minipage} &
    \begin{minipage}{.22\textwidth}
      \includegraphics[width=\linewidth]{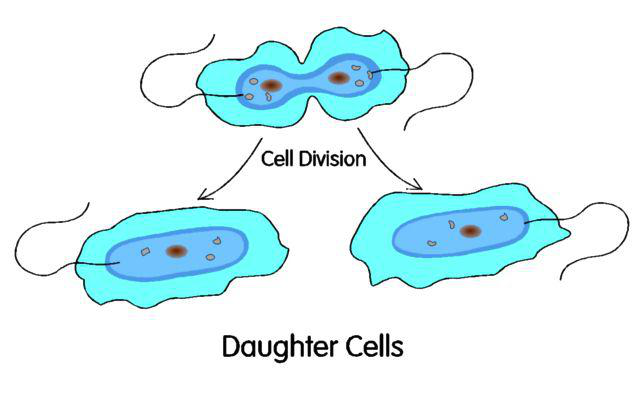}
    \end{minipage} &
    \begin{minipage}{.22\textwidth}
      \includegraphics[width=\linewidth]{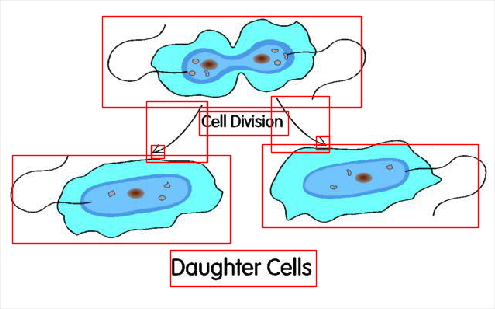}
    \end{minipage} &
    \begin{minipage}{.23\textwidth}
      \includegraphics[width=\linewidth]{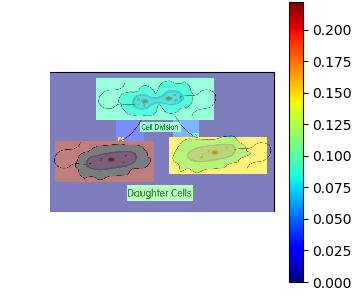}
    \end{minipage} 
    \\ 
    \begin{minipage}{.21\textwidth}
       Which lamps would turn on if switch is connected?
    \vspace{-3mm}
    \begin{enumerate}[noitemsep]
     \item b \checkmark
    \item a
    \item a, b, c
    \item c
    \end{enumerate}
    \end{minipage} &
    \begin{minipage}{.22\textwidth}
      \includegraphics[width=\linewidth]{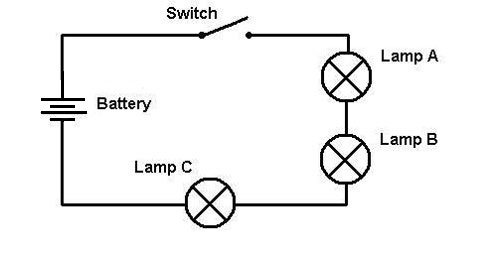}
    \end{minipage} &
    \begin{minipage}{.22\textwidth}
      \includegraphics[width=\linewidth]{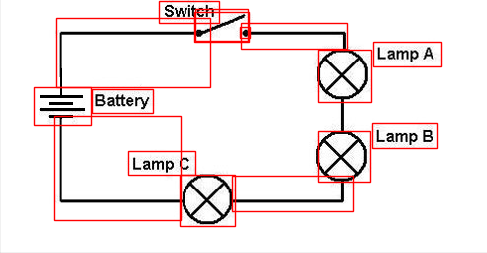}
    \end{minipage} &
    \begin{minipage}{.23\textwidth}
      \includegraphics[width=\linewidth]{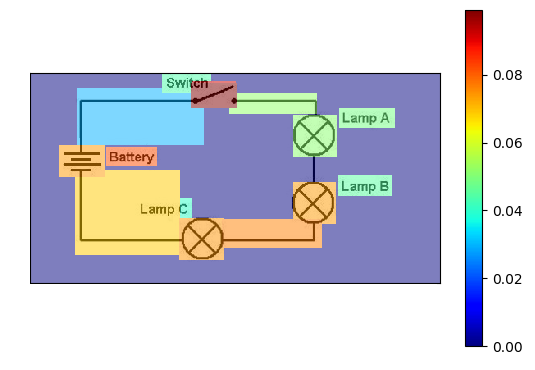}
    \end{minipage} 
    \\ 
    \begin{minipage}{.21\textwidth}
         In which state does the substance hold shape?
    \vspace{-3mm}
    \begin{enumerate}[noitemsep]
    \item solid \checkmark
    \item liquid
    \item gas
    \item none
    \end{enumerate}
    \end{minipage} &
    \begin{minipage}{.23\textwidth}
      \includegraphics[width=\linewidth]{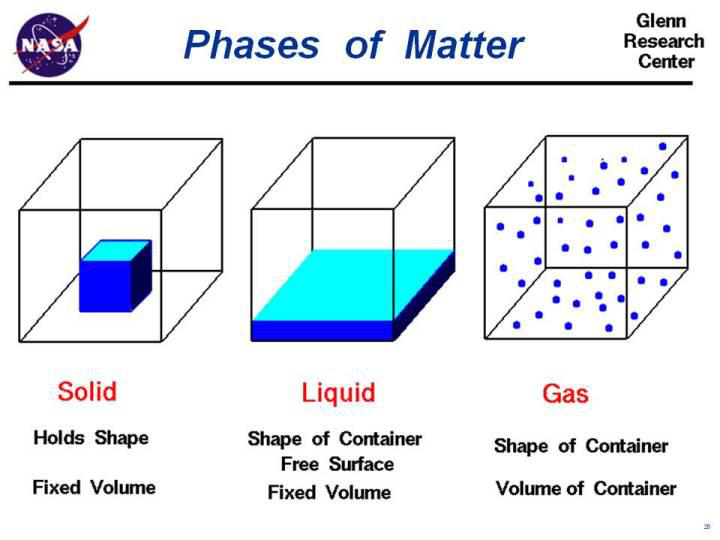}
    \end{minipage} &
    \begin{minipage}{.22\textwidth}
      \includegraphics[width=\linewidth]{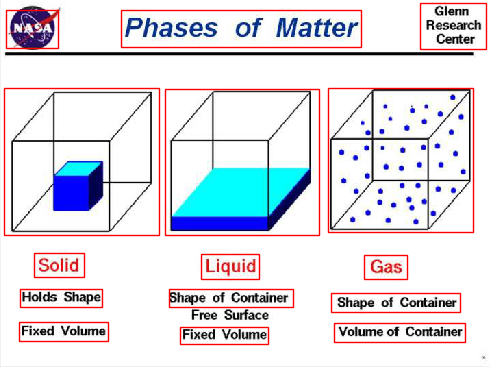}
    \end{minipage} &
    \begin{minipage}{.22\textwidth}
      \includegraphics[width=\linewidth]{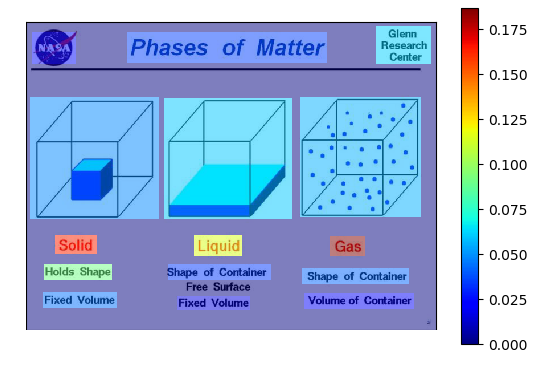}
    \end{minipage} \\ 
    \bottomrule
  \end{tabular}
  \caption{Study of the attention on question diagrams. Examples (earth, life sciences, physics) from validation set.}
  \label{table:qual}
\end{table*}
Table~\ref{table:qual} illustrates the impact of the different levels of attention on the question diagram, previously quantified in table~\ref{table:DMC}. The first column shows question and answer options, while the second adds the question diagram. The third illustrates the RoIs extracted through BU attention and the fourth adds attention heatmaps using BUTD. While BU assigns equal attention to all RoIs, BUTD also attends to the text of the question and each possible answer. For all the example questions, only the model with BUTD produces the correct answer.

Some observations follow. RoI detection  (column three) fails to detect inner shapes in nested diagrams, e.g. state of matter, question four. The heatmaps in column four show a rather low maximum BUTD attention intensity, between 0.08 and 0.2, which may suggest an opportunity to improve the cross-modal aspects of our model. When the text of the correct answer (\textit{topsoil}, \textit{solid}) appears explicitly in the diagram, its RoI is generally more attended than the rest. Other times, the RoI labeled by such text (\textit{switch}) is the warmest. ISAAQ seems to attend to aspects of the diagrams that are key to answer correctly, suggesting both language and visual understanding: the two cells resulting from the original one through mitosis, the circuit segment where a lamp flashes upon switch actuation or the properties of the state in which a substance holds shape. Other examples seem to indicate some ability to deal with counting and spatial reasoning. All will deserve further investigation in future work.

\section{Conclusion}
\label{sec:conc}
This paper reports on ISAAQ, the first system to achieve accuracies above 80\%, 70\% and 55\% on TQA true/false, text and diagram MC questions. ISAAQ demonstrates that it is possible to master the grand AI challenge of machine textbook understanding based on modern methods for language and visual understanding, with modest infrastructure requirements. Key to this success are transformers, BUTD attention, pre-training on related datasets, and the selection of complementary background information to train and ensemble different solvers. Our approach allowed overcoming critical challenges like the complexity and relatively small size of the TQA dataset or the scarcity of large diagram datasets. Still, further research is necessary to keep pushing the boundaries of textbook understanding, e.g. by charting and expanding the reasoning skills of transformers, making model outcomes more interpretable by humans, and further exploiting diagrams. Additional effort will also be needed in activities like the development of large diagram datasets, including the semantic annotation of diagram constituents and connectors, and annotating diagram questions with the reasoning and knowledge types required to answer them. 

\section*{Acknowledgements}
This research was funded by the Horizon 2020 grant European Language Grid-825627. Special gratitude to Peter Clark and the Allen Institute for sharing data and insights on the ARC datasets.


\bibliography{emnlp2020}

\begin{thebibliography}{39}
\expandafter\ifx\csname natexlab\endcsname\relax\def\natexlab#1{#1}\fi

\bibitem[{{Anderson} et~al.(2018){Anderson}, {He}, {Buehler}, {Teney},
  {Johnson}, {Gould}, and {Zhang}}]{8578734}
Peter {Anderson}, Xiaodong {He}, Chris {Buehler}, Damien {Teney}, Mark
  {Johnson}, Stephen {Gould}, and Lei {Zhang}. 2018.
\newblock Bottom-up and top-down attention for image captioning and visual
  question answering.
\newblock In \emph{2018 IEEE/CVF Conference on Computer Vision and Pattern
  Recognition}, pages 6077--6086.

\bibitem[{Antol et~al.(2015)Antol, Agrawal, Lu, Mitchell, Batra, Zitnick, and
  Parikh}]{Antol2015VQAVQ}
Stanislaw Antol, Aishwarya Agrawal, Jiasen Lu, Margaret Mitchell, Dhruv Batra,
  C.~Lawrence Zitnick, and Devi Parikh. 2015.
\newblock Vqa: Visual question answering.
\newblock \emph{2015 IEEE International Conference on Computer Vision (ICCV)},
  pages 2425--2433.

\bibitem[{Boratko et~al.(2018)Boratko, Padigela, Mikkilineni, Yuvraj, Das,
  McCallum, Chang, Fokoue-Nkoutche, Kapanipathi, Mattei, Musa, Talamadupula,
  and Witbrock}]{boratko-etal-2018-systematic}
Michael Boratko, Harshit Padigela, Divyendra Mikkilineni, Pritish Yuvraj,
  Rajarshi Das, Andrew McCallum, Maria Chang, Achille Fokoue-Nkoutche, Pavan
  Kapanipathi, Nicholas Mattei, Ryan Musa, Kartik Talamadupula, and Michael
  Witbrock. 2018.
\newblock \href {https://doi.org/10.18653/v1/W18-2607} {A systematic
  classification of knowledge, reasoning, and context within the {ARC}
  dataset}.
\newblock In \emph{Proceedings of the Workshop on Machine Reading for Question
  Answering}, pages 60--70, Melbourne, Australia. Association for Computational
  Linguistics.

\bibitem[{Clark et~al.(2018)Clark, Cowhey, Etzioni, Khot, Sabharwal, Schoenick,
  and Tafjord}]{Clark2018ThinkYH}
Peter Clark, Isaac Cowhey, Oren Etzioni, Tushar Khot, Ashish Sabharwal, Carissa
  Schoenick, and Oyvind Tafjord. 2018.
\newblock Think you have solved question answering? try arc, the ai2 reasoning
  challenge.
\newblock \emph{ArXiv}, abs/1803.05457.

\bibitem[{Clark et~al.(2019)Clark, Etzioni, Khashabi, Khot, Mishra, Richardson,
  Sabharwal, Schoenick, Tafjord, Tandon, Bhakthavatsalam, Groeneveld, Guerquin,
  and Schmitz}]{Clark2019FromT}
Peter Clark, Oren Etzioni, Daniel Khashabi, Tushar Khot, Bhavana~Dalvi Mishra,
  Kyle Richardson, Ashish Sabharwal, Carissa Schoenick, Oyvind Tafjord, Niket
  Tandon, Sumithra Bhakthavatsalam, Dirk Groeneveld, Michal Guerquin, and
  Michael Schmitz. 2019.
\newblock From 'f' to 'a' on the n.y. regents science exams: An overview of the
  aristo project.
\newblock \emph{ArXiv}, abs/1909.01958.

\bibitem[{Clark et~al.(2020)Clark, Tafjord, and
  Richardson}]{Clark2020TransformersAS}
Peter Clark, Oyvind Tafjord, and Kyle Richardson. 2020.
\newblock Transformers as soft reasoners over language.
\newblock \emph{ArXiv}, abs/2002.05867.

\bibitem[{Dauphin et~al.(2017)Dauphin, Fan, Auli, and
  Grangier}]{10.5555/3305381.3305478}
Yann~N. Dauphin, Angela Fan, Michael Auli, and David Grangier. 2017.
\newblock Language modeling with gated convolutional networks.
\newblock In \emph{Proceedings of the 34th International Conference on Machine
  Learning - Volume 70}, ICML'17, page 933–941. JMLR.org.

\bibitem[{Denaux and Gomez-Perez(2019)}]{Denaux2019Vecsigrafo}
Ronald Denaux and Jose~Manuel Gomez-Perez. 2019.
\newblock \href {https://doi.org/10.3233/SW-190361} {Vecsigrafo: Corpus-based
  word-concept embeddings}.
\newblock \emph{Semantic Web}, pages 1--28.

\bibitem[{Deng et~al.(2009)Deng, Dong, Socher, Li, Li, and
  Fei-Fei}]{imagenet_cvpr09}
J.~Deng, W.~Dong, R.~Socher, L.-J. Li, K.~Li, and L.~Fei-Fei. 2009.
\newblock {ImageNet: A Large-Scale Hierarchical Image Database}.
\newblock In \emph{CVPR09}.

\bibitem[{Devlin et~al.(2018)Devlin, Chang, Lee, and Toutanova}]{Devlin2018}
Jacob Devlin, Ming{-}Wei Chang, Kenton Lee, and Kristina Toutanova. 2018.
\newblock \href {http://arxiv.org/abs/1810.04805} {{BERT:} pre-training of deep
  bidirectional transformers for language understanding}.
\newblock \emph{CoRR}, abs/1810.04805.

\bibitem[{Gomez-Perez and Ortega(2019)}]{gomezlre2019}
Jose~Manuel Gomez-Perez and Raul Ortega. 2019.
\newblock \href {https://doi.org/10.1145/3360901.3364420} {Look, read and
  enrich - learning from scientific figures and their captions}.
\newblock In \emph{Proceedings of the 10th International Conference on
  Knowledge Capture}, K-CAP ’19, page 101–108, New York, NY, USA.
  Association for Computing Machinery.

\bibitem[{{He} et~al.(2016){He}, {Zhang}, {Ren}, and {Sun}}]{resnet2016}
K.~{He}, X.~{Zhang}, S.~{Ren}, and J.~{Sun}. 2016.
\newblock Deep residual learning for image recognition.
\newblock In \emph{2016 IEEE Conference on Computer Vision and Pattern
  Recognition (CVPR)}, pages 770--778.

\bibitem[{Howard and Ruder(2018)}]{howard2018ulmfit}
Jeremy Howard and Sebastian Ruder. 2018.
\newblock \href {http://arxiv.org/abs/1801.06146} {Fine-tuned language models
  for text classification}.
\newblock \emph{CoRR}, abs/1801.06146.

\bibitem[{Kembhavi et~al.(2016)Kembhavi, Salvato, Kolve, Seo, Hajishirzi, and
  Farhadi}]{Kembhavi2016ADI}
Aniruddha Kembhavi, Mike Salvato, Eric Kolve, Minjoon Seo, Hannaneh Hajishirzi,
  and Ali Farhadi. 2016.
\newblock A diagram is worth a dozen images.
\newblock In \emph{ECCV}.

\bibitem[{Kembhavi et~al.(2017)Kembhavi, Seo, Schwenk, Choi, Farhadi, and
  Hajishirzi}]{Kembhavi2017AreYS}
Aniruddha Kembhavi, Minjoon Seo, Dustin Schwenk, Jonghyun Choi, Ali Farhadi,
  and Hannaneh Hajishirzi. 2017.
\newblock Are you smarter than a sixth grader? textbook question answering for
  multimodal machine comprehension.
\newblock \emph{2017 IEEE Conference on Computer Vision and Pattern Recognition
  (CVPR)}, pages 5376--5384.

\bibitem[{Khashabi et~al.(2020)Khashabi, Khot, Sabharwal, Tafjord, Clark, and
  Hajishirzi}]{Khashabi2020UnifiedQACF}
Daniel Khashabi, Tushar Khot, Ashish Sabharwal, Oyvind Tafjord, P.~Clark, and
  Hannaneh Hajishirzi. 2020.
\newblock Unifiedqa: Crossing format boundaries with a single qa system.
\newblock \emph{ArXiv}, abs/2005.00700.

\bibitem[{Kim et~al.(2019)Kim, Kim, and Kwak}]{Kim2019TextbookQA}
DaeSik Kim, Seonhoon Kim, and Nojun Kwak. 2019.
\newblock Textbook question answering with multi-modal context graph
  understanding and self-supervised open-set comprehension.
\newblock In \emph{ACL}.

\bibitem[{Kingma and Ba(2014)}]{Kingma2014AdamAM}
Diederik~P. Kingma and Jimmy Ba. 2014.
\newblock Adam: A method for stochastic optimization.
\newblock \emph{CoRR}, abs/1412.6980.

\bibitem[{Krishna et~al.(2017)Krishna, Zhu, Groth, Johnson, Hata, Kravitz,
  Chen, Kalantidis, Li, Shamma, Bernstein, and
  Fei-Fei}]{10.1007/s11263-016-0981-7}
Ranjay Krishna, Yuke Zhu, Oliver Groth, Justin Johnson, Kenji Hata, Joshua
  Kravitz, Stephanie Chen, Yannis Kalantidis, Li-Jia Li, David~A. Shamma,
  Michael~S. Bernstein, and Li~Fei-Fei. 2017.
\newblock \href {https://doi.org/10.1007/s11263-016-0981-7} {Visual genome:
  Connecting language and vision using crowdsourced dense image annotations}.
\newblock \emph{Int. J. Comput. Vision}, 123(1):32–73.

\bibitem[{{Li} et~al.(2018){Li}, {Su}, {Zhu}, {Wang}, and {Zhang}}]{li2018}
Juzheng {Li}, Hang {Su}, Jun {Zhu}, Siyu {Wang}, and Bo~{Zhang}. 2018.
\newblock Textbook question answering under instructor guidance with memory
  networks.
\newblock In \emph{2018 IEEE/CVF Conference on Computer Vision and Pattern
  Recognition}, pages 3655--3663.

\bibitem[{Lin et~al.(2014)Lin, Maire, Belongie, Bourdev, Girshick, Hays,
  Perona, Ramanan, Doll{\'a}r, and Zitnick}]{Lin2014MicrosoftCC}
Tsung-Yi Lin, Michael Maire, Serge Belongie, Lubomir Bourdev, Ross Girshick,
  James Hays, Pietro Perona, Deva Ramanan, Piotr Doll{\'a}r, and C.~Lawrence
  Zitnick. 2014.
\newblock Microsoft coco: Common objects in context.
\newblock In \emph{ECCV}.

\bibitem[{Liu et~al.(2019)Liu, Ott, Goyal, Du, Joshi, Chen, Levy, Lewis,
  Zettlemoyer, and Stoyanov}]{Liu2019RoBERTaAR}
Yinhan Liu, Myle Ott, Naman Goyal, Jingfei Du, Mandar Joshi, Danqi Chen, Omer
  Levy, Mike Lewis, Luke Zettlemoyer, and Veselin Stoyanov. 2019.
\newblock Roberta: A robustly optimized bert pretraining approach.
\newblock \emph{ArXiv}, abs/1907.11692.

\bibitem[{Mihaylov et~al.(2018)Mihaylov, Clark, Khot, and
  Sabharwal}]{mihaylov-etal-2018-suit}
Todor Mihaylov, Peter Clark, Tushar Khot, and Ashish Sabharwal. 2018.
\newblock \href {https://doi.org/10.18653/v1/D18-1260} {Can a suit of armor
  conduct electricity? a new dataset for open book question answering}.
\newblock In \emph{Proceedings of the 2018 Conference on Empirical Methods in
  Natural Language Processing}, pages 2381--2391, Brussels, Belgium.
  Association for Computational Linguistics.

\bibitem[{Peters et~al.(2018)Peters, Neumann, Iyyer, Gardner, Clark, Lee, and
  Zettlemoyer}]{Peters}
Matthew Peters, Mark Neumann, Mohit Iyyer, Matt Gardner, Christopher Clark,
  Kenton Lee, and Luke Zettlemoyer. 2018.
\newblock \href {https://doi.org/10.18653/v1/N18-1202} {Deep contextualized
  word representations}.
\newblock In \emph{Proceedings of the 2018 Conference of the North {A}merican
  Chapter of the Association for Computational Linguistics: Human Language
  Technologies, Volume 1 (Long Papers)}, pages 2227--2237, New Orleans,
  Louisiana. Association for Computational Linguistics.

\bibitem[{Petroni et~al.(2019)Petroni, Rockt{\"a}schel, Riedel, Lewis, Bakhtin,
  Wu, and Miller}]{petroni-etal-2019-language}
Fabio Petroni, Tim Rockt{\"a}schel, Sebastian Riedel, Patrick Lewis, Anton
  Bakhtin, Yuxiang Wu, and Alexander Miller. 2019.
\newblock \href {https://doi.org/10.18653/v1/D19-1250} {Language models as
  knowledge bases?}
\newblock In \emph{Proceedings of the 2019 Conference on Empirical Methods in
  Natural Language Processing and the 9th International Joint Conference on
  Natural Language Processing (EMNLP-IJCNLP)}, pages 2463--2473, Hong Kong,
  China. Association for Computational Linguistics.

\bibitem[{Radford et~al.(2018)Radford, Narasimhan, Salimans, and
  Sutskever}]{radford2018OpenAITransformer}
Alec Radford, Karthik Narasimhan, Tim Salimans, and Ilya Sutskever. 2018.
\newblock Improving language understanding by generative pre-training.
\newblock \emph{URL https://s3-us-west-2. amazonaws.
  com/openai-assets/research-covers/languageunsupervised/language understanding
  paper. pdf}.

\bibitem[{Reddy(1988)}]{Reddy_1988}
Raj Reddy. 1988.
\newblock \href {https://doi.org/10.1609/aimag.v9i4.950} {Foundations and grand
  challenges of artificial intelligence: Aaai presidential address}.
\newblock \emph{AI Magazine}, 9(4):9.

\bibitem[{Redmon et~al.(2015)Redmon, Divvala, Girshick, and
  Farhadi}]{Redmon2015YouOL}
Joseph Redmon, Santosh~Kumar Divvala, Ross~B. Girshick, and Ali Farhadi. 2015.
\newblock You only look once: Unified, real-time object detection.
\newblock \emph{2016 IEEE Conference on Computer Vision and Pattern Recognition
  (CVPR)}, pages 779--788.

\bibitem[{Ren et~al.(2017)Ren, He, Girshick, and
  Sun}]{10.1109/TPAMI.2016.2577031}
Shaoqing Ren, Kaiming He, Ross Girshick, and Jian Sun. 2017.
\newblock \href {https://doi.org/10.1109/TPAMI.2016.2577031} {Faster r-cnn:
  Towards real-time object detection with region proposal networks}.
\newblock \emph{IEEE Trans. Pattern Anal. Mach. Intell.}, 39(6):1137–1149.

\bibitem[{Seo et~al.(2017)Seo, Kembhavi, Farhadi, and
  Hajishirzi}]{Seo2017BidirectionalAF}
Minjoon Seo, Aniruddha Kembhavi, Ali Farhadi, and Hannaneh Hajishirzi. 2017.
\newblock Bidirectional attention flow for machine comprehension.
\newblock \emph{CoRR}, abs/1611.01603.

\bibitem[{Sharma et~al.(2018)Sharma, Ding, Goodman, and
  Soricut}]{sharma-etal-2018-conceptual}
Piyush Sharma, Nan Ding, Sebastian Goodman, and Radu Soricut. 2018.
\newblock \href {https://doi.org/10.18653/v1/P18-1238} {Conceptual captions: A
  cleaned, hypernymed, image alt-text dataset for automatic image captioning}.
\newblock In \emph{Proceedings of the 56th Annual Meeting of the Association
  for Computational Linguistics (Volume 1: Long Papers)}, pages 2556--2565,
  Melbourne, Australia. Association for Computational Linguistics.

\bibitem[{Shoeybi et~al.(2019)Shoeybi, Patwary, Puri, LeGresley, Casper, and
  Catanzaro}]{Shoeybi2019MegatronLMTM}
M.~Shoeybi, M.~Patwary, R.~Puri, P.~LeGresley, J.~Casper, and Bryan Catanzaro.
  2019.
\newblock Megatron-lm: Training multi-billion parameter language models using
  model parallelism.
\newblock \emph{ArXiv}, abs/1909.08053.

\bibitem[{Simonyan and Zisserman(2015)}]{DBLP:journals/corr/SimonyanZ14a}
Karen Simonyan and Andrew Zisserman. 2015.
\newblock \href {http://arxiv.org/abs/1409.1556} {Very deep convolutional
  networks for large-scale image recognition}.
\newblock In \emph{3rd International Conference on Learning Representations,
  {ICLR} 2015, San Diego, CA, USA, May 7-9, 2015, Conference Track
  Proceedings}.

\bibitem[{Su et~al.(2019)Su, Zhu, Cao, Li, Lu, Wei, and Dai}]{su2019vl}
Weijie Su, Xizhou Zhu, Yue Cao, Bin Li, Lewei Lu, Furu Wei, and Jifeng Dai.
  2019.
\newblock Vl-bert: Pre-training of generic visual-linguistic representations.
\newblock \emph{arXiv preprint arXiv:1908.08530}.

\bibitem[{Sun et~al.(2019)Sun, Yu, Yu, and Cardie}]{sun-etal-2019-improving}
Kai Sun, Dian Yu, Dong Yu, and Claire Cardie. 2019.
\newblock \href {https://doi.org/10.18653/v1/N19-1270} {Improving machine
  reading comprehension with general reading strategies}.
\newblock In \emph{Proceedings of the 2019 Conference of the North {A}merican
  Chapter of the Association for Computational Linguistics: Human Language
  Technologies, Volume 1 (Long and Short Papers)}, pages 2633--2643,
  Minneapolis, Minnesota. Association for Computational Linguistics.

\bibitem[{Tan and Bansal(2019)}]{tan-bansal-2019-lxmert}
Hao Tan and Mohit Bansal. 2019.
\newblock \href {https://doi.org/10.18653/v1/D19-1514} {{LXMERT}: Learning
  cross-modality encoder representations from transformers}.
\newblock In \emph{Proceedings of the 2019 Conference on Empirical Methods in
  Natural Language Processing and the 9th International Joint Conference on
  Natural Language Processing (EMNLP-IJCNLP)}, pages 5100--5111, Hong Kong,
  China. Association for Computational Linguistics.

\bibitem[{Teney et~al.(2017)Teney, Liu, and
  Hengel}]{Teney2017GraphStructuredRF}
Damien Teney, L.~Liu, and A.~V.~D. Hengel. 2017.
\newblock Graph-structured representations for visual question answering.
\newblock \emph{2017 IEEE Conference on Computer Vision and Pattern Recognition
  (CVPR)}, pages 3233--3241.

\bibitem[{Weston et~al.(2014)Weston, Chopra, and Bordes}]{Weston2014MemoryN}
Jason Weston, Sumit Chopra, and Antoine Bordes. 2014.
\newblock Memory networks.
\newblock \emph{CoRR}, abs/1410.3916.

\bibitem[{Zhu et~al.(2015)Zhu, Kiros, Zemel, Salakhutdinov, Urtasun, Torralba,
  and Fidler}]{10.1109/ICCV.2015.11}
Yukun Zhu, Ryan Kiros, Rich Zemel, Ruslan Salakhutdinov, Raquel Urtasun,
  Antonio Torralba, and Sanja Fidler. 2015.
\newblock \href {https://doi.org/10.1109/ICCV.2015.11} {Aligning books and
  movies: Towards story-like visual explanations by watching movies and reading
  books}.
\newblock In \emph{Proceedings of the 2015 IEEE International Conference on
  Computer Vision (ICCV)}, ICCV ’15, page 19–27, USA. IEEE Computer
  Society.

\end{thebibliography}
\bibliographystyle{acl_natbib}

\end{document}